# ON THE PERFORMANCE OF FILTERS FOR REDUCTION OF SPECKLE NOISE IN SAR IMAGES OFF THE COAST OF THE GULF OF GUINEA

KlogoGriffith S.[1], GasonooAkpeko[2] and Ampomah K. E. Isaac[3]

[1,2,3] Department of Computer Engineering Kwame Nkrumah University of Science and Technology, KNUST, Kumasi, Ghana

## ABSTRACT

*Synthetic Aperture Radar (SAR) imagery to monitor oil spills are some methods that have been proposed for the West African sub-region. With the increase in the number of oil exploration companies in Ghana (and her neighbors) and the rise in the coastal activities in the sub-region, there is the need for proper monitoring of the environmental impact of these socio-economic activities on the environment.Detection and near real-time information about oil spills are fundamental in reducing oil spill environmental impact. SAR images are prone to some noise, which is predominantly speckle noise around the coastal areas. This paper evaluatesthe performance of the mean and median filters used in the preprocessing filtering to reduce speckle noise in SAR images for most image processing algorithms.*

## 1. INTRODUCTION

The West African sub-region is one of the active oil exploration regions in the world, with countries such as Nigeria, Ghana, Ivory Coast, Democratic Republic of Congo Cameroon and Equatorial Guinea been the major producers. The Gulf of Guinea countries are estimated to produce about 4% of the global total of oil per day, with Nigeria producing more than half of the total for the sub-region [1]. Ghana as a country is not new to oil exploration, the country's oil exploration dates back to about a century in the saltpond fields.Ghana discovered oil in commercial quantities in June 2007. However, with the increase in the number of oil exploration companies in Ghana (and her neighbors) and the rise in the coastal activities in the sub-region, there is the need for proper monitoring of the environmental impact of these socio-economic activities on the environment. The increase in the exploration activities in the sub-region as shown in figure 1and figure 2 comes with its attendant effects of oil spills which are usually intentional or accidental.

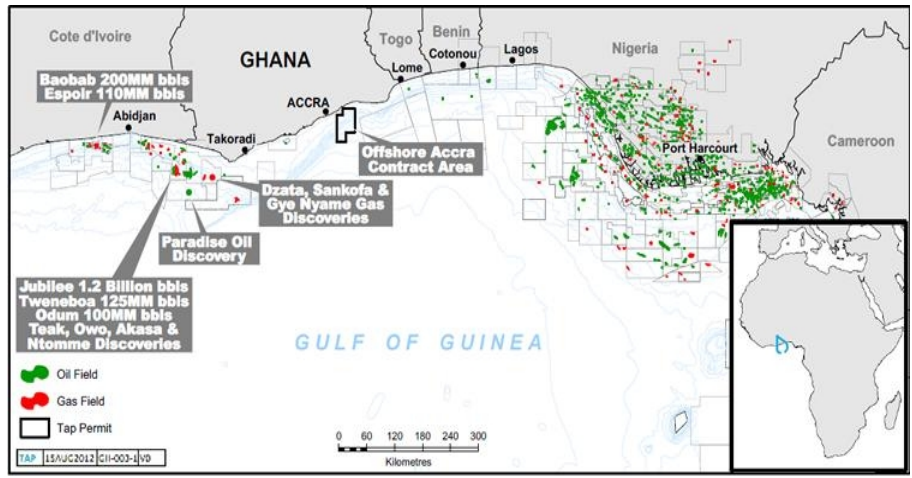

Fig.1: Oil Exploration in the Gulf of Guinea [2]

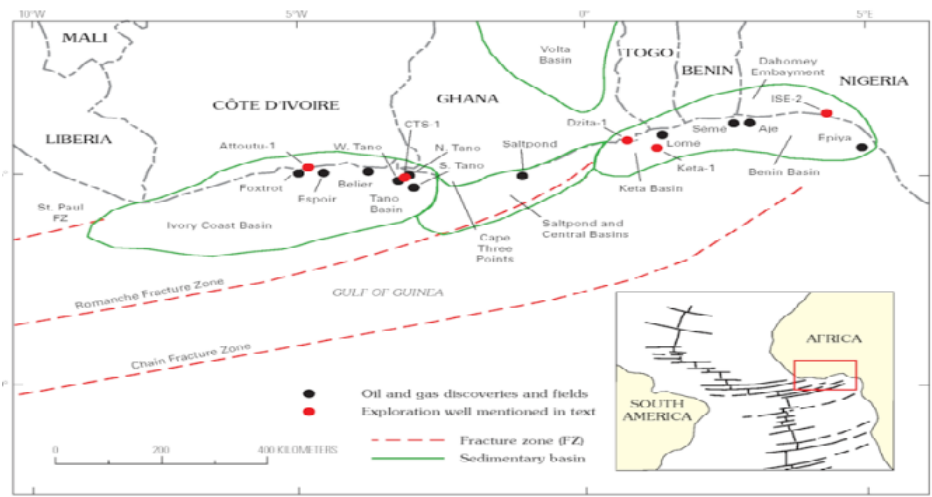

Fig.2: Discovery and Exploration fields in The Gulf of Guinea [3]

Oil spills can come from various sources including leakage from oil transportation tankers, accidents on oil platforms and seepage from natural seeps.Oil spills are getting more and more frequent on the sea surface. Annually, 48% of the oil pollution in the seas is caused by fuels and 29% by crude oil. Tanker accidents contribute with only 5% of the all pollution entering into the sea [4]. Timely, accurate and continuous detection of oil spills in the sub-region can help management and monitoring of oil spills.Remote sensing does provides fast and reliable measurement of the ocean's surface waters, and thus has been used widely to assess oil pollution at sea, as well as other water bodies. The remote sensing instruments include optical, microwave, and radar (e.g., Synthetic Aperture Radar, SAR) sensors which can be airborne and satellites.Airborne remote sensing offers the highest resolution and fastest response in measurements, but often very expensive to be used for operational monitoring. In contrast, satellite remote sensing provides reliable measurements of the ocean at a reasonable cost, hence more suitable for oil spill monitoring [5]. SAR is frequently used for oil spill detection at sea, compared to the various satellite instruments. The sea-surface echo radar signal is modulated by the wind-induced capillary waves, and thus carries detailed information on the surface roughness [6], giving the degree of oil spill on the surface of the sea. Due to the difference in the surface tensions of water and oil, there is also a difference in the backscattered signals that is received by the instrument.Oil film on the ocean surface has higher surface tension than water, and doesreduce the surface capillary waves, resulting in a smoother sea surface. The irregularity of the radar echo signal makes the oil contaminated areas appear as dark patches in SAR images.SAR is an all-weather and all-day high resolution aerial and space imaging of terrain. Being independent of light and weather conditions, SAR images does perform better than optical images hence its efficiency for detecting various ocean surface features such as oil spills. The backscattered or radar echo energy level for oil spilled areas received in the SAR systems is too low since the oil reduces the capillary waves of the sea surface. However, other natural phenomenon also contributes to damping the short waves and creates dark areas of the sea surface in the SAR images. Some of these natural phenomenons that create dark areas on the sea surface are due to suspension of Bragg scattering mechanism depending on ocean and/or atmospheric conditions [6]. Atmospheric conditions also contribute to noise in the SAR images, which affects the interpretation and/or feature extraction from the images. Signals received by radar are usually contaminated by noise due to random modulation of the radar pulse during atmospheric propagation, or due to fluctuation in the receiving circuits.

## 2. NOISE AND NOISE REDUCTION

Noise is a random and usually an unwanted signal in a lot of applications: it can be experienced in acoustics signals, picture and video signals. Noise is evident in the variation in brightness or color information in a picture or a video sequence. Noise is most obvious in regions with low signal level, such as the weak received echo-signal in a radar receiver. Noise is characterized by its statistical properties. Noise containing all frequencies with equal amplitudes called "white" noise is typical in picture and video applications. Radar images are often corrupted by these random variations in intensity values. Some common types of noise are salt and pepper noise, impulse noise, and Gaussian noise.Some of the noise that affects SAR images includes Speckle Noise and Gaussian Noise. To better perceive images, noise reduction techniques need to be applied to reduce the unwanted areas of the image.

### 2.1. Speckle Noise

Speckle noise is a coarse noise that is usually evident in and degrades the quality of the active radar and synthetic aperture radar (SAR) images. Radar waves can interfere constructively or destructively to produce light and dark pixels in radar image. Speckle noise is commonly observed in radar sensing system and images, although it can be observed in most types of remotely sensed images utilizing coherent radiation. Speckle noise in radar data or images have multiplicative error and must be removed before the data can be used otherwise the noise is merged into and degrades the image quality.

## 3. FILTERS

Broadly filters are designed to remove unwanted components from a set of signals, they can be devices or processes. In image processing, filters are used to improve the quality of images prior to its usage. Filtering is also sometimes referred to as smoothing, which is done to reduce noise and improve the image quality.Depending on the type of noise, linear or nonlinear filters are employed to remove the noise. Linear filters are good filters for removing Gaussian noise andother types of noise in most cases as well. Linear filters are implemented using the weighted sum of the pixels in successive windows of the image. Nonlinear filters are those that are implemented without a weighted sum of pixels. Nonlinear filters are usually spatially invariant, which implies that the same calculation is performed at all parts of the image. The Mean and Median filters are typical linear and nonlinear filters employed in image processing respectively.

### 3.1. Mean filter

Mean or Average filter is one of the simplest linear filters that is implemented by performing local averaging operation as shown in equation 1 and the value of each image pixel is replaced by the average of all the values in the local neighborhood [7][8]. The mean filter works as low-pass one.

$$h[i,j] = \frac{1}{M} \sum_{(k,l) \in N} f[k,l] \qquad 1$$

where *M* is the number of pixels in the surrounding *N*, *h[i,j]* and *f[k,l]* are the old and new image pixels.The size of the neighborhood *N* controls the degree of filtering. A large neighborhood size will result in a greater degree of filtering.

### 3.2. Median filter

The median filter is a nonlinear filter usually spatially invariant, which replaces each pixel value with the median of the pixel values in the local neighborhood [7]. The median filter is very effective in retaining the image details since they do not depend on values which are significantly different from typical values in the neighborhood. The median filter works on consecutive image window in a manner similar to that of the linear filters, but the method employed does not use a weighted sum.Pixels in each window are sorted into ascending order and the pixel value in the middle is selected as the new value for a particular pixel.

## 4. PERFORMANCE METRICS

Performance of filters can be estimated using numerical measures of picture quality after the filter has been applied to the image. The performance metrics are chosen based on their computable distortion measures.Mean squared error (MSE) and Peak Signal to Noise Ratio (PSNR) are by far the most common measures of picture quality in image systems [9] [10].

### 4.1. Mean-Square Error (MSE)

The easiest and widely used image quality metric is the mean squared error (MSE), computed by averaging the squared intensity differences of distorted and reference image pixels. MSE

International Journal of Information Technology, Modeling and Computing (IJITMC) Vol.1,No.4,November 2013

application adopts an assumption that the reduction of perceptual quality of an image is directlyrelated to the visibility of the error signal. The image signal whose quality is being evaluated is thought of as a sum of an accurate reference signal and an error signal [10].MSE can objectively quantify the strength of the error signal present in an image after a filter is applied to the same image. The mean squared error (MSE) for practical purposes, allows comparison of the pixel values of imagesafter filtering to the degraded image before filtering.

### 4.2. Peak Signal-to-Noise Ratio (PSNR)

Peak signal-to-noise ratio (PSNR) of an image is a relation for the ratio between the maximum value (power) of a signal and the power of distorting noise that disturbs the quality of its image [11]. Peak signal-to-noise ratio (PSNR) is a related quality metric which is usually used along with MSE. These two are appealing due to the simplicity to calculate, easy physical meanings and mathematically convenient in the context of optimization [10].

## 5. EVALUATION OF MEAN AND MEDIAN FILTERS ON SPECKLE NOISE

The mean and median filters are compared to evaluate their performance in the reduction of speckle noise in SAR images. The filters are evaluated using image pixel window size of *3X3* in MATLAB, while introducing various percentage of speckle noise into the image.The original image in RGB is shown in figure 3 and the gray level of the image with the histogram distribution is shown in figure 4 and 5 respectively. The image with 50% speckle noise with the histogram distribution is shown in figure 6 and 7 respectively.

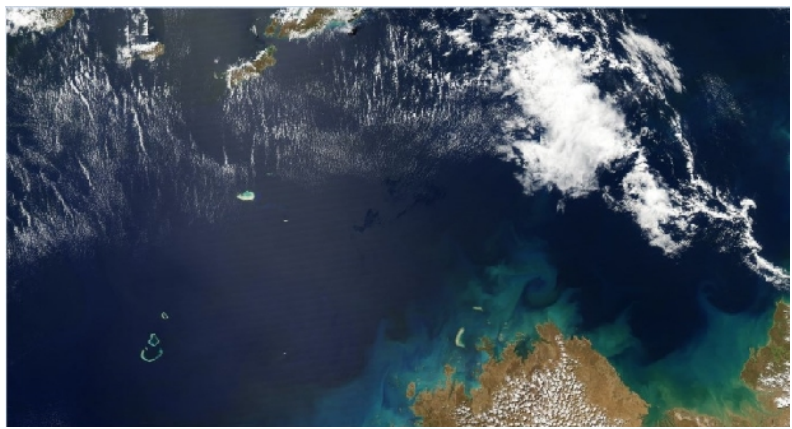

Fig. 3: SAR image (RGB) without speckle noise

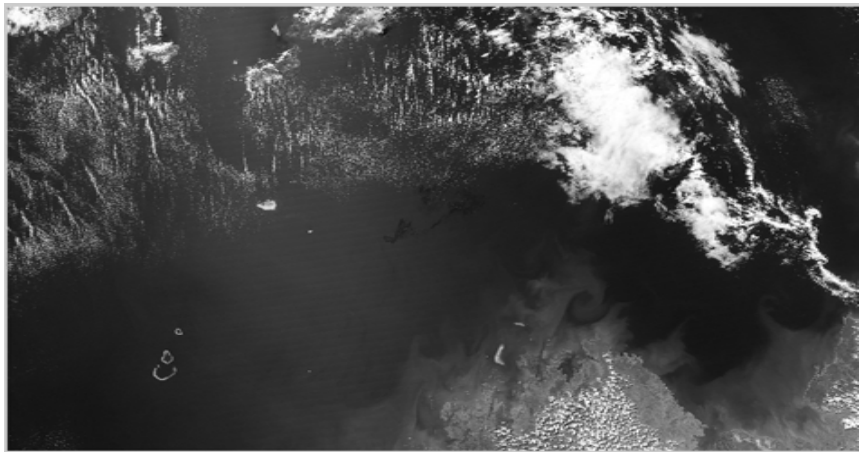

Fig. 4: SAR image (gray) without speckle noise

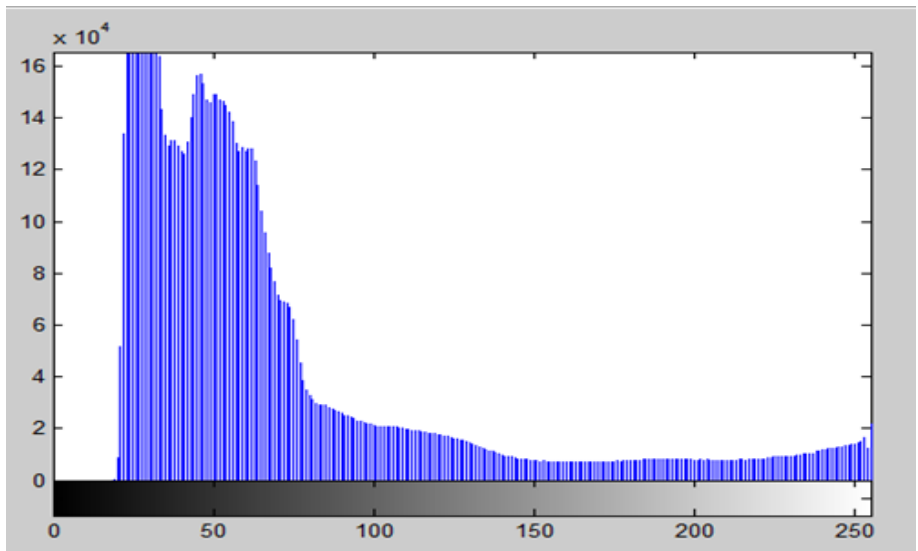

Fig. 5: Histogram distribution of SAR image (gray) without speckle noise

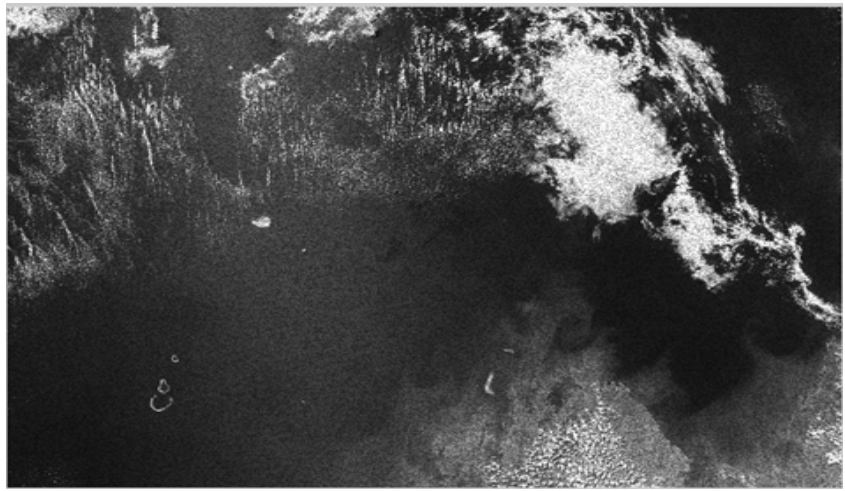

Fig. 6: SAR image (gray) with speckle noise

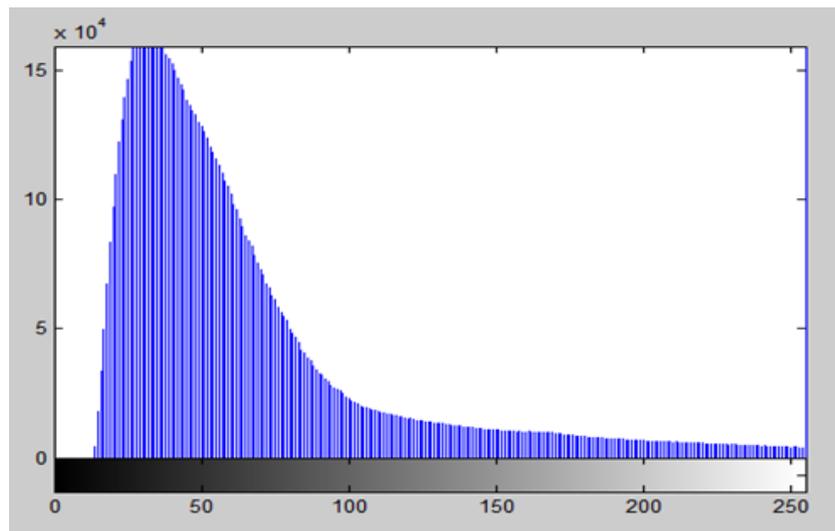

Fig. 7: Histogram distribution of SAR image (gray) with speckle noise

## 6. PERFORMANCE OF MEAN AND MEDIAN FILTERS ON SPECKLE NOISE

In comparing the performance of the mean and median filters on the reduction of speckle noise, MSE and PSNR was used.Table 1 and 2 summarizes the performance of the mean and median filters for the various levels of speckle noise. Figure 8 and 9 are graphical simulation showing the performance of the mean and median filters with reference to MSE and PSNR respectively.

Table 1: Comparison of PSNR for mean and median filters

| Filter type | Percentage of speckle noise in image ||||||||||||||||||
|---|---|---|---|---|---|---|---|---|---|---|---|---|---|---|---|---|---|---|
| | 0.01 | 0.02 | 0.03 | 0.04 | 0.05 | 0.06 | 0.07 | 0.08 | 0.09 | 0.1 | 0.2 | 0.3 | 0.4 | 0.5 | 0.6 | 0.7 | 0.8 | 0.9 |
| | PSNR (Peak Signal to noise ratio) in decibel (dB) ||||||||||||||||||
| Mean Filter | 28.65 | 26.52 | 25.14 | 24.11 | 23.29 | 22.62 | 22.04 | 21.53 | 21.09 | 20.69 | 18.00 | 16.42 | 15.34 | 14.67 | 14.18 | 13.80 | 13.49 | 13.23 |
| Median Filter | 24.57 | 23.46 | 22.60 | 21.90 | 21.30 | 20.78 | 20.32 | 19.92 | 19.55 | 19.21 | 16.85 | 15.39 | 14.38 | 13.72 | 13.25 | 12.87 | 12.57 | 12.30 |

Table 2: Comparison of MSE for mean and median filters

| Filter type | Percentage of speckle noise in image ||||||||||||||||||
|---|---|---|---|---|---|---|---|---|---|---|---|---|---|---|---|---|---|---|
| | 0.01 | 0.02 | 0.03 | 0.04 | 0.05 | 0.06 | 0.07 | 0.08 | 0.09 | 0.1 | 0.2 | 0.3 | 0.4 | 0.5 | 0.6 | 0.7 | 0.8 | 0.9 |
| | Mean-Square Error (MSE) ||||||||||||||||||
| Mean Filter | 14.79 | 18.53 | 21.44 | 23.89 | 25.99 | 27.95 | 29.70 | 31.39 | 32.92 | 34.32 | 45.99 | 54.15 | 59.93 | 62.46 | 63.38 | 63.61 | 63.53 | 63.10 |
| Median Filter | 25.24 | 26.93 | 28.29 | 29.41 | 30.49 | 31.48 | 32.37 | 33.23 | 33.95 | 34.71 | 40.92 | 45.59 | 49.57 | 52.88 | 55.74 | 58.41 | 60.83 | 62.86 |

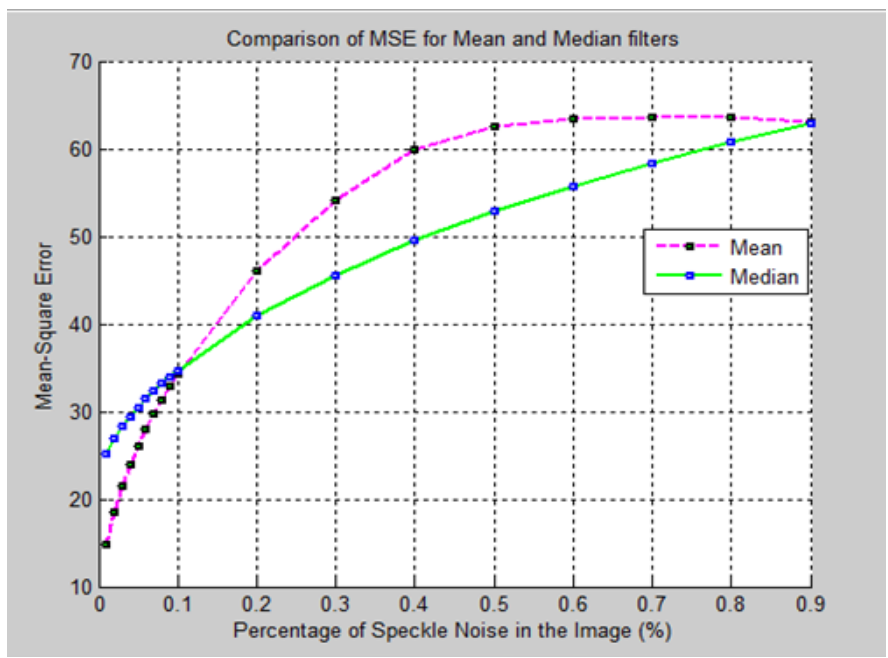

Fig. 8: Graph of comparison of MSE for mean and median filters

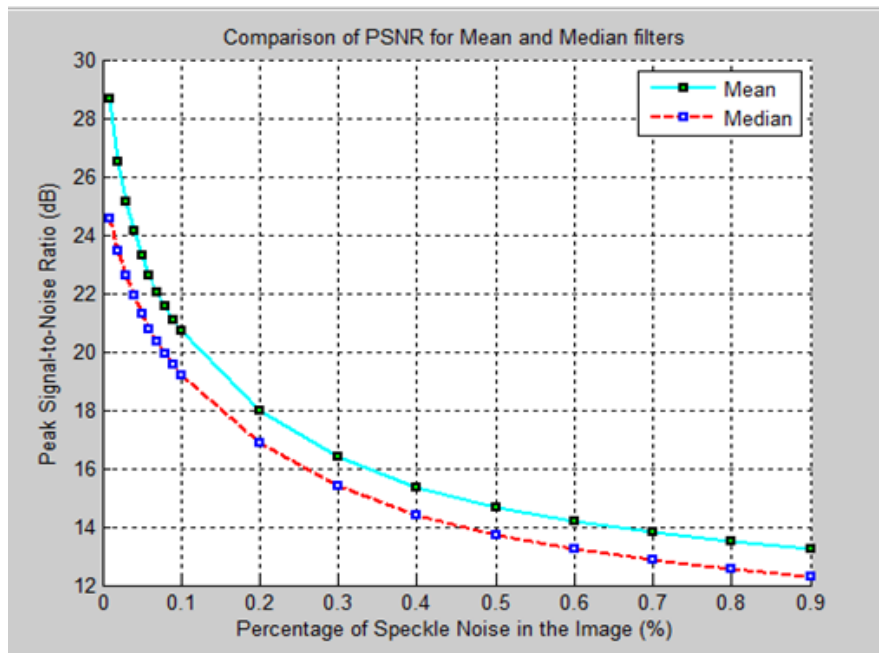

Fig. 9: Graph of comparison of PSNR for mean and median filters

## 7. DISCUSSION OF RESULTS

In the comparison of the performance of the mean and median filters on the reduction of speckle noise in SAR images, various degrees of speckle noise was introduced into the image from 1% to 90%. Using an image pixel window size of *3X3*, the results for MSE show that the mean filter performs better than the median filter when the noise level is below 10% as shown in figure 8. The median filter performs better than the mean filter for high levels of noise for both the MSE and PSNR as shown in figure 8 and 9.

## 8. CONCLUSION

In this paper, the mean and median filters were compared to evaluate their performance in reduction of speckle noise typical in radar images like SAR images. The simulation results show that the median filters performs better for high levels of speckle noise in the SAR image. The mean filter is performs very well in terms of MSE when the noise levels are low. Therefore, designing an adaptive algorithm which will take advantage of the strengths of both the mean and median filters will be a possible future research direction.

International Journal of Information Technology, Modeling and Computing (IJITMC) Vol.1,No.4,November 2013